\documentclass{article}

\usepackage{amsmath}
\usepackage{amsfonts}
\usepackage{amssymb}
\usepackage[title]{appendix}
\usepackage{algorithm}
\usepackage[noend]{algpseudocode}
\usepackage{graphicx}
\usepackage{url}

\usepackage[preprint]{neurips_2020}
\usepackage[utf8]{inputenc} 
\usepackage[T1]{fontenc}    
\usepackage{hyperref}       
\usepackage{url}            
\usepackage{booktabs}       
\usepackage{amsfonts}       
\usepackage{nicefrac}       
\usepackage{microtype}      

\author{Michael Tetelman\\
BayzAI, Volkswagen Group of America, Innovation Center California\\
\texttt{michael.tetelman@gmail.com}}
\title{On Compression Principle and Bayesian Optimization for Neural Networks
}
\begin{document}

\maketitle

\begin{abstract}
Finding methods for making generalizable predictions is a fundamental problem of machine learning. By looking into similarities between the  prediction problem for unknown data and the lossless compression we have found an approach that gives a solution. In this paper we propose a compression principle that states that an optimal predictive model is the one that minimizes a total compressed message length of all data and model definition while guarantees decodability. Following the compression principle we use Bayesian approach to build probabilistic models of data and network definitions. A method to approximate Bayesian integrals using a sequence of variational approximations is implemented as an optimizer for hyper-parameters: Bayesian Stochastic Gradient Descent (BSGD). Training with BSGD is completely defined by setting only three  parameters: number of epochs, the size of the dataset and the size of the minibatch, which define a learning rate and a number of iterations. We show that dropout can be  used for a continuous dimensionality reduction that allows to find optimal network dimensions as required by the compression principle.
\end{abstract}

\section{Introduction}

One of the most intriguing properties of neural networks models is their ability to learn from available data samples and make predictions about new data, see \citep{lecun2015deep}.

The predictions for new data cannot be absolutely accurate and the prediction errors give a very important characteristic of the neural network model.

Making a prediction about some vector value $x$ means to compute a probability $P(x)$. Unless specifically stated here and throughout the paper $x$ includes all data sample components which for predictive models typically are pairs $x=(input, labels) \equiv (X,Y)$. We will consider models that predict both input and label components, so the probabilistic model is a chain of submodels for input and label parts $P(x) \equiv P(X,Y) = P(Y|X)P(X)$.

The definition of a neural network model provides a specific recipe - an algorithm for computing the probability of $x$ that depends on an exact definition of the network architecture $arch$ and a set of numerical parameters commonly called weights $w$ \citep{goodfellow2016deep}

\begin{equation}\label{model:1}
P(x|w,arch) \equiv P(x|w).
\end{equation}

To simplify notations we will omit $arch$ argument when it is possible without effecting the meaning.

Prediction error or loss for a known ground truth value $x$ is defined as 

\begin{equation}\label{loss}
l(x,w) = -\log{P(x|w)}.
\end{equation}

While it is possible to minimize the loss for a known $x$ by selecting an appropriate network definition and network weights the prediction error for a new data sample which is unknown at the time of prediction will be most of the time higher than for optimized loss.

The novelty gap between the prediction loss for unknown and known data points is inevitable and reflects the information gap between unknown and available data.
The goal of the predictive method is to find an algorithm that is expected to minimize that gap. 

We would like to note that the prediction problem is very much similar to the problem of a lossless compression of data. The lossless compression problem typically set as follows: there is a sender that creates an encoded message about data points and sends it to a receiver who decodes the message and finds what data are. Like in the prediction problem a sender's goal is to minimize a message length that is given by negative log of probability of the model in use \eqref{loss} and send enough information so the receiver will be able to decode the message \citep{mackay2003information}.

The decodability of the compressed message is a very important constraint. To decode the message the receiver must know the definition of the model  \eqref{model:1} used by the sender. If the sender will use the weights $w_0$ that minimizes the message length \eqref{loss} it is necessary also to send an additional message describing $w_0$ by using a probabilistic model of weights $P(w)$ that must be known by a receiver.
Then the total message length will be 

\begin{equation}\label{totalloss}
l_{tot}(x,w_0) = -\log{P(x|w_0)} - \log{P(w_0)}.
\end{equation}

The eq. \eqref{totalloss} above shows that the optimal weights correspond to a minimum of the total message length for both data and weight descriptions and not the minimum of the message length for the data sample alone.

The encoded message for weights must be send to a receiver because it is required for decodability. Essentially the extra message length is equivalent (but not equal) to a novelty gap for predicting new data.
We consider that equivalence in \eqref{totalloss} to be more than a coincidence and rather as a fundamental property.

We propose here \emph{the compression principle} that states that an optimal predictive model is the one that minimizes a total compressed message length of all data and model definition while guarantees decodability.

The message length of the model works as a regularizer that allows to avoid overfitting: if model has a big number of weights we can find the weight values that will reduce a compressed data length to be  very small, however, the compressed model size will be larger and so the total length.

Another consequence of \emph{the compression principle} is that the optimal model size will increase with the size of available data. In particular this is an important outcome of the neural architecture search (NAS) based on \emph{the compression principle} - this is not in a scope of the current paper and will be considered elsewhere.

The contributions of the paper are follows: 

\begin{enumerate}
\item A Bayesian stochastic gradient descent optimizer for hyper-parameters (BSGD) is derived by using \emph{the compression principle} applied to a neural network optimization;
\item A method for computing Bayesian integrals is developed based on  renormalization group ideas by computing differentials of the hyper-parameters via sequential variational approximations;
\item We show that dropout effectively works as a dimensionality reduction method, that allows to continuously control an effective number of network parameters.
\end{enumerate}

The paper is organized as follows: in section \ref{compression} we discuss the compression approach, in section \ref{dropout} dropout as a dimensionality reduction tool, in section \ref{integrals} the method for approximate computing of Bayesian integrals; in section \ref{BSGD} we derive an optimizer - Bayesian SGD, examples of using BSGD are discussed in section \ref{experiments}.

\section{The Compression Principle}
\label{compression}

We will consider data samples as randomly and independently drawn from an unknown source distribution.

The probability of a dataset $\{x_n;n=1..N\}$ derived from a defined model $P(x|w)$ is given by the following Bayesian integral over  weights multiplied by corresponding prior of hyper-parameters and network architecture
\begin{equation}\label{compression:integral}
 P(\{x_n\},\mathbf{H},arch) = P(\mathbf{H}|arch) P(arch) \int\displaylimits_{w}
 {P(w|\mathbf{H},arch)\prod_{n=1}^{N}{P(x_n|w)} \mathrm{d}w}, 
\end{equation} 
where $P(w|\mathbf{H},arch)$ is a prior distribution of weights that depends on hyper-parameters $\mathbf{H}$ and network architecture $arch$, $P(\mathbf{H}|arch)$ is a prior distribution of hyper-parameters for a given architecture and $P(arch)$ is a prior of the network architecture.

To make a prediction about data point $x_0$ we need to compute a ratio of two Bayesian integrals

\begin{equation}\label{compression:prediction}
P(x_0|\{x_n\},\mathbf{H},arch) = P(\{x_0, x_n\},\mathbf{H},arch)/P(\{x_n\},\mathbf{H},arch)
\end{equation}

To achieve a better compression we need to maximize the probability of dataset w.r.t. hyper-parameters $\mathbf{H}$ and architecture. To simplify notations further we will include architecture of the network into hyper-parameter variables $\mathbf{H}$.

\emph{The compression principle} requires decodability, which means that the compressed message should provide a complete description of  all hyper-parameters including network architecture that allows to reconstruct both weight prior $P(w|\mathbf{H})$ and model $P(x|w)$. The decodability results in an expanded message length by the additional part that describes all hyper-parameters

\begin{equation}\label{compression:length}
l_{total} = l_{data} + l_{\mathbf{H}}; \;
l_{\mathbf{H}} = -\log{P(\mathbf{H}|arch)} -\log{P(arch)}.
\end{equation}

Reducing the number of parameters helps to avoid overfitting. However there are other methods known to improve the predictive performance like dropout without explicitly reducing dimensions of parameters \citep{dropout2014}.

We show here that dropout effectively works as a dimensionality reduction method.

\section{Dropout as a dimensionality reduction}
\label{dropout}

Dropout layer is implemented by randomly replacing feature values with zeros with some rate $r, 0\leqslant r < 1$. Due to dropout the information content of a feature vector is reduced. To account for that let's consider a representation of features as a binary string with one bit per feature. Each bit has value 1 if the feature is positive and zero otherwise. The information content in the bit-string is compatible with the definition of ReLU activations, which are essentially binary gate units that transfer input as is if it is positive and output zero otherwise. 

Dropout noise is reducing channel capacity for each bit in the bit-string of features which in the presence of noise is given by a mutual information $I(X;Y)$ between one bit of input $X$ and one bit of output $Y$ of the dropout layer:

\begin{equation}
I(X;Y) = \sum_{X,Y}{p(X)p(Y|X)\log_2{\left(\frac{p(Y|X)}{p(Y)}\right)}},
\end{equation}

where $p(Y|X=0)=\delta_{Y,0}$ and 
$p(Y|X=1)=r\delta_{Y,0}+(1-r)\delta_{Y,1}$.
Considering most informative input distribution $p(X=0)=p(X=1)=1/2$ dropout with rate $r$ reduces one bit of information to a fraction of it

\begin{equation}\label{dropout:dimred}
I(X;Y) = 1 - 0.5\left\lbrace{r\log_2{\frac{1}{r}} + (1+r)\log_2{(1+r)}}\right\rbrace.
\end{equation}

See details for general case in the Appendix \ref{appendix:dropout}. For example, the dropout rate $r=0.5$ in eq.\eqref{dropout:dimred} results in a  factor 0.3113, which reduces the effective number of features to almost $1/3$ of the original number. Then all effective dimensions of weight matrices should be adjusted by multiplying by  square of the corresponding reduction factors, which for $r=0.5$ reduces the effective number of weights by the factor $0.0969 \approx 0.1$.

\section{Computing Bayesian integrals}
\label{integrals}

For approximate computing of the Bayesian integrals in the eq. \eqref{compression:integral} we will use an idea of renormalization group combined with the variational inference \citep{mackay2003information}.

The integral over weights is taken from a product of the prior of the weight distribution and a product of a large number of model probabilities for each data sample that can be represented as an exponential of the negative loss for a dataset of $N$ samples

\begin{equation}\label{int:int}
I(\{x_n\},\mathbf{H}) = \int{\mathrm{d}w P(w|\mathbf{H})\prod_{n=1}^N P(x_n|w)} = \int{\mathrm{d}w P(w|\mathbf{H}) e^{-L(w)}},
L = -\sum_{n=1}^N{l_n(w)}.
\end{equation}

The renormalization group approach here is based on using the same parametrization for the posterior as for the prior, so to approximate the posterior distribution at the initial hyper-parameter point  $\mathbf{H_0}$ we use a prior distribution at a different hyper-parameter point $\mathbf{H_1}$

\begin{equation}
P(w|\mathbf{H_0}) e^{-L(w)} \sim P(w|\mathbf{H_1}).
\end{equation}

To achieve better accuracy we will do it incrementally in a sequence of small steps as follows: we split a small part of loss $L(w)=\epsilon L + (1-\epsilon L)$ with a small $\epsilon \ll 1$ and factorize the integral in eq.\eqref{int:int} as follows

\begin{equation}\label{int:factor}
I(\{x_n\},\mathbf{H_0}) = \mathbf{\langle AB \rangle_{H_1}} \approx \mathbf{\langle A \rangle_{H_1} \langle B \rangle_{H_1}}
\end{equation}

where $\mathbf{A}$ and $\mathbf{B}$ are

\begin{equation}\label{int:ab}
\mathbf{A} = \frac{P(w|\mathbf{H_0})e^{-\epsilon L}}{
P(w|\mathbf{H_1})}, \; \; \mathbf{B} = e^{-(1-\epsilon)L}.
\end{equation}

and the average of some $\mathbf{C}$ is defined as 
$\mathbf{\langle C \rangle_{H_1}} = \int{\mathrm{d}w P(w|\mathbf{H_1}) \mathbf{C}(w)}$ where $\mathbf{C}$ could be $\mathbf{A, B}$ or $\mathbf{AB}$.

The error of replacing the average of a product with a product of averages can be estimated by using the Cauchy-Schwarz inequality \citep{cauchySchwarz} in the following form

\begin{equation}\label{cauchy-schwarz}
\left(\mathbf{\langle AB \rangle_{H_1} - \langle A \rangle_{H_1} \langle B \rangle_{H_1}}\right)^2 \leqslant 
\mathbf{Var\left(A\right) Var\left(B\right)}.
\end{equation}

The error in eq.\eqref{int:factor} is minimized by selecting hyper-parameters $\mathbf{H_1}$ at a minimum of the variance of $\mathbf{A}$ which gives the following update rule:

\begin{equation}\label{int:update}
\mathbf{H_1} = \mathbf{H_0} - 
\epsilon \frac{1}{\langle\left(\frac{\partial\log{P(w|\mathbf{H_0})}}{\partial\mathbf{H}}\right)^2\rangle_{\mathbf{H_0}}} 
\frac{\partial \langle L \rangle_{\mathbf{H_0}}}{\partial\mathbf{H}}
\end{equation}

where 

\begin{equation}
\langle L \rangle_{\mathbf{H_0}} = \int{\mathrm{d}w P(w|\mathbf{H_0}) L(w)},
\end{equation}
\begin{equation}
\langle \left(\frac{\partial\log{P(w|\mathbf{H_0})}}{\partial\mathbf{H}}\right)^2 \rangle_{\mathbf{H_0}} = 
\int{\mathrm{d}w P(w|\mathbf{H_0})\left(\frac{\partial\log{P(w|\mathbf{H_0})}}{\partial\mathbf{H}}\right)^2}.
\end{equation}

The eq.\eqref{int:update} describes a gradient descent in the hyper-parameter space.
The variance of $\mathbf{A}$ at a minimal point $\mathbf{H_1}$ has an order of value $O(\epsilon^2)$ and goes to zero with $\epsilon \rightarrow 0$ and so the error of the approximation of the integral for each step goes to zero as $\epsilon$. Taking into account the independence of $N$ random samples we can find that for the one step integral approximation the error is $O(\epsilon \sqrt{N})$. See details in the Appendix \ref{appendix:integral}.

Now, after the first step the integral in eq.\eqref{int:int} is approximately equal to

\begin{equation}
I(\{x_n\},\mathbf{H_0}) \approx \\
\int{\mathrm{d}w P(w|\mathbf{H_0}) e^{-\epsilon L}}
\int{\mathrm{d}w P(w|\mathbf{H_1}) e^{-(1-\epsilon) L}}.
\end{equation}

The whole integral is computed by repeating these steps exactly $T$ times using the update rule \eqref{int:update} with $\mathbf{H_0, H_1}$ replaced by $\mathbf{H_t, H_{t+1}}$ for all  $t=0..(T-1)$, where $T=1/\epsilon$ is a number of epochs. Then the integral is equal to

\begin{equation}\label{int:total}
I(\{x_n\},\mathbf{H_0}) = \prod_{t=0}^{T-1}
\left( \int{\mathrm{d}w P(w|\mathbf{H_t}) e^{-\epsilon L(w)}} \right).
\end{equation}

The posterior distribution is given by $P(w|\mathbf{H}_{T-1})$ where $\mathbf{H}_{T-1}$ is a final value of hyper-parameters from recurrent updates in eq.\eqref{int:update}.

The total error of approximating the integral \eqref{int:int} after $T$ epochs accumulates to an exponential factor $\exp(O(\sqrt{N}))$ which for large $N$ gives a small correction relative to the value of the integral $I \propto \exp(O({N}))$ in the eq.\ref{int:total}.

\section{Bayesian SGD}
\label{BSGD}

The developed method for an approximate computation of the Bayesian integrals gives us a general approach for the hyper-parameter optimization. Unlike the approaches based on heuristics and searching  the hyper-parameter space \citep{yu2020hyperparameter} our method by derivation completely determines the optimal parameter values like learning rate and number of iterations for the solution.

In the previous section we considered a method of computing the Bayesian integrals with a sequence of steps where each step accounts for a contribution of a fraction of the total loss that corresponds to  a full batch gradient descent. Here we are considering another multi-step method where we are using for each step a fraction of a loss for a single randomly selected data sample or a minibatch $l_n(w)$.
For the minibatch $l_n$ is a normalized loss per sample. The update rule will become then a stochastic gradient descent in the  hyper-parameter space

\begin{equation}\label{bsgd-update}
\mathbf{H_{t+1}} = \mathbf{H_t} - 
\epsilon \frac{b}{\langle\left(\frac{\partial\log{P(w|\mathbf{H_t})}}{\partial\mathbf{H}}\right)^2\rangle_{\mathbf{H_t}}} 
\frac{\partial \langle l_n \rangle_{\mathbf{H_t}}}{\partial\mathbf{H}},
\end{equation}

where $b$ is a minibatch size.

Total number of the iterations needed to complete computing the integral $T$ is equal to a number of epochs $N_e$ times a number of minibatches in the dataset $N_b$: $T=N_e N_b$, while parameter $\epsilon=1/N_e$.

The eq.\eqref{bsgd-update} defines a general hyper-parameter optimization method based on Bayesian SGD. For the following we will consider an important practical case of a fixed architecture with Gaussian prior distribution of $d$ independent weights where hyper-parameters are means and variances $(\mu_i, \sigma_i), i=1..d$

\begin{equation}
P(w|\mathbf{H}) = \prod_{i=1}^d{\frac{e^{-\frac{(w_i-\mu_i)^2}{2\sigma_i^2}}}
{\sqrt{2\pi\sigma_i^2}}}, \quad \mathbf{H} = \{(\mu_i,\sigma_i); i=1..d\}.
\end{equation}

While in the eq.\eqref{bsgd-update} the factor with a square of the the gradient contains a direct product and actually is a matrix by hyper-parameters we will simplify it by using only diagonal elements of the matrix. Then it is easy to find that for each $i=1..d$

\begin{equation}
\langle\left(\frac{\partial\log{P(w|\mathbf{\mu,\sigma)}}}{\partial\mu_i}\right)^2\rangle = \frac{1}{\sigma_i^2}, \quad
\langle\left(\frac{\partial\log{P(w|\mathbf{\mu,\sigma)}}}{\partial\sigma_i}\right)^2\rangle = \frac{2}{\sigma_i^2}
\end{equation}

and update rules will look as follows

\begin{equation}\label{bsgd-mu-sigma}
\mu_{i,t+1} = \mu_{i,t} - \epsilon b\; \sigma_{i,t}^2
\frac{\partial\langle l_n\rangle_t}{\partial \mu_{i,t}}, \quad
\sigma_{i,t+1} = \sigma_{i,t} - \epsilon b\; \frac{\sigma_{i,t}^2}{2}
\frac{\partial\langle l_n\rangle_t}{\partial \sigma_{i,t}}.
\end{equation}

The gradients of the Gaussian average of the minibatch loss could be transformed to averages of the gradients w.r.t weights

\begin{equation}
\frac{\partial\langle l_n\rangle_t}{\partial \mu_{i,t}} = 
\langle\frac{\partial l_n}{\partial w_{i,t}}\rangle_t, \quad
\frac{\partial\langle l_n\rangle_t}{\partial \sigma_{i,t}} = 
\sigma_{i,t}\langle\frac{\partial^2 l_n}{\partial w_{i,t}^2}\rangle_t.
\end{equation}

Let's define a scaled inverse variance $s_i = 1/(\sigma_i^2 b)$, where $b$ is the minibatch size with total number of samples $N=bN_b$. 
Then finally, the equations for Bayesian SGD have the following form

\begin{equation}\label{bsgd:final}
\mu_{i,t+1} = \mu_{i,t} - \frac{\epsilon}{s_{i,t}}
\langle\frac{\partial l_n}{\partial w_{i,t}}\rangle_t, \quad
s_{i,t+1} = s_{i,t} + \epsilon\;
\langle\frac{\partial^2 l_n}{\partial w_{i,t}^2}\rangle_t,
\end{equation}

here the learning rate is $\epsilon=1/N_e$ with the total number of iterations $T=N_e N_b$.

In practical computing the second gradient by weights is replaced by a square of the fist gradient by weights $\partial^2 l/\partial w^2 \rightarrow (\partial l/\partial w)^2$. Analysis of that equivalence is given in the Appendix \ref{appendix:bsgd}.

Before a first iteration all inverse variances $s$ are initialized with ones and all means $\mu$ are initialized with a standard weight initialization. Then training is performed for exactly $T$ iterations. Hyper-parameters $(\mu,\sigma)$ are updated at each step as in Algorithm \ref{algo:bsgd}.

\begin{algorithm}
\caption{Bayesian SGD for one step}\label{algo:bsgd}
\begin{algorithmic}[1]
\State select minibatch index: $n$
\State get weight sample from normal distribution: $w \gets \mathcal{N}(\mu, 1/{\sqrt{sb}})$
\State compute the minibatch loss per sample: $l_n(w)$
\State compute gradient of the loss: $\mathbf{grad_n} \gets {\partial l_n}/{\partial w}$
\State compute update for $\mu$: $\mu \gets \mu - \frac{1}{N_e} \mathbf{grad_n}/s$
\State compute update for $s$: $s \gets s +\frac{1}{N_e}(\mathbf{grad_n})^2$
\end{algorithmic}
\end{algorithm}

The sampling of weights from the weight prior is needed to approximate the averaging of losses by the weight prior. We should note that the weight sampling that was proposed earlier was based on heuristics: the \emph{dropconnect} in \citep{dropconnect} and the \emph{Gaussian dropout} in \citep{pmlr-v28-wang13a}. In \citep{kingma2015variational} the variational dropout was proposed as  generalization of the Gaussian dropout.

The Bayesian SGD has only three free parameters: a number of epochs $N_e$, a number of minibatches in a training set $N_b$ and the size of the minibatch that together define a total number of iterations and a learning rate.

\section{Experiments}
\label{experiments}

The experiments were implemented in Pytorch for MNIST classification problem \citep{mnist}. Data were used as is without augmentation, preprocessing or pre-training. The convolutional network was trained on a standard training set of 60K images with 28x28 pixels per an image with three different optimizers: BSGD, ADAM \citep{kingma2014adam}, SGD \citep{ruder2016overview} with the same training conditions: minibatch size 60, training length 10 epochs, with total number of iterations 10K. Training was done on a single GPU machine, with a total training time for  any optimizer about 10 min. Testing and validation were computed with a standard 10K-sample testing set. No validation or testing results were used in training.

The learning rate for ADAM was set to $0.0001$ while learning rate for SGD was set to $0.1$. Learning rate for BSGD is set automatically to the inverse number of epoch which is $0.1$. Dropout was used for all non-linear layers with the dropout rate $0.01$.

The standard cross-entropy loss was used for training with any  optimizer.

The network consists of 26 layers total: 1 input convolutional layer, 9 dual convolutional residual blocks, adaptive pooling, 3 fully connected dual residual blocks and a final linear layer. The detailed network definition can be found in Appendix \ref{appendix:net}.

\begin{figure}
\begin{center}
\includegraphics[scale=0.555]{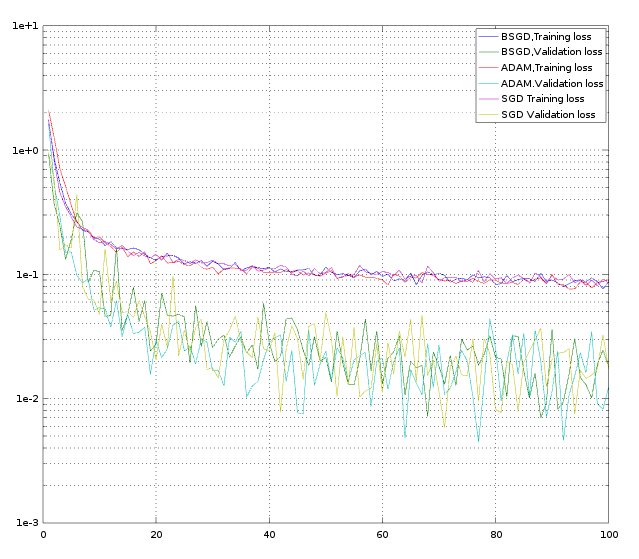}
\end{center}
	
\caption{Progress of training and validation losses for BSGD, ADAM, SGD given in log scale vs number of epochs. Every 10 points on horizontal axis corresponds to one epoch with 10 epochs total.}
	
\label{fig:t-v-loss}
\end{figure}

There were no significant differences found in a training progress for the train and validation losses per sample between all 3 optimizers as shown in Figure. \ref{fig:t-v-loss}.

However, the prediction accuracy on the testing set was notably different, with the maximum accuracy over multiple runs 0.9975 obtained by BSGD. Average accuracy over multiple runs for each optimizer is given in the Table \ref{table:accuracy}.

\begin{table}[t!]
\centering
\caption{Average accuracy for each tested optimizer.}
\begin{tabular}{|c c c|}
\hline
BSGD & ADAM & SGD \\
\hline
0.996323 & 0.994642 & 0.99298 \\
\hline
\end{tabular}
\label{table:accuracy}
\end{table}

The performance for ADAM and SGD optimizers depends on specific values of a number of parameters and first of all on the learning rate. The used learning rate $10^{-4}$ for ADAM and $0.1$ for SGD was selected by trial and error. There is a possibility that a better performance could be achieved with a different selection of the learning rate and other optimizer parameters. However, there is no method known  other than exhaustive search for finding the optimal parameters in this case. In practical computing that results in a significant time spent on tuning optimizer parameters via multiple trials.

Bayesian SGD does not have that problem. In fact it has the highest performance when all parameters in the Algorithm \ref{algo:bsgd} have values that derived from approximating the Bayesian integrals and lower performance for different values.

\begin{figure}
\begin{center}
\includegraphics[scale=0.555]{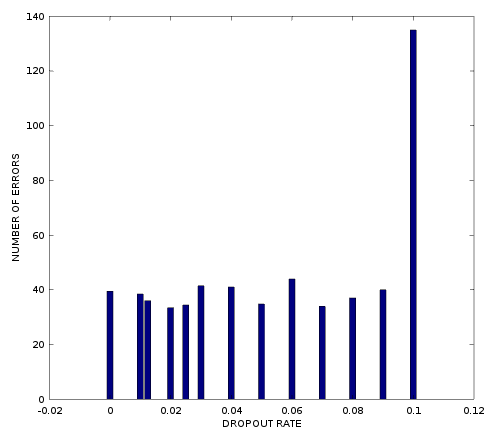}
\end{center}
\caption{Average number of errors vs dropout rate for the same network}
\label{fig:error-dropout}
\end{figure}

Interesting to see an effect of dropout on the accuracy. Figure \ref{fig:error-dropout} shows that an average number of test errors  does not change  significantly for a large interval of dropout rates from 0 to 0.09. The errors were averaged over multiple training runs of the same network definition for the same dropout rate.

On the other side using the estimate of the effective number of feature dimensions \ref{dropout:dimred} we can find that the effective number of parameters of the network is reduced to 0.6 of the original size. That allows to pose a question "What is an actual number of parameters that parametrize a solution?"

\section{Conclusion}

We developed a method for an approximate computing of the posterior given by the Bayesian integral over weights via iterative updates of the hyper-parameters using variational approximations and reparameterization of the prior to represent the posterior. 

We derived the Bayesian SGD optimizer from a fundamental Bayesian approach. The BSGD algorithm is automatically defines an optimal learning rate per hyper-parameter and a number of iterations.

We tested BSGD, ADAM, and SGD on a standard MNIST classification problem by training with equal conditions for each optimizer. After averaging over multiple runs the BSGD shows the best performance.

We studied the effect of dropout on the test performance of the networks trained with BSGD optimizer and found no significant difference in the performance for an interval of dropout rates from 0 to 0.09.

\section*{Broader Impact}

We think that the proposed \emph{compression principle} will attract an interest of the research community.

We think that the developed method for estimating the Bayesian integrals will be found useful for researchers in the field of Deep Learning.

We think that the proposed Bayesian SGD optimizer will allow researchers and practitioners to significantly speed up training neural networks and finding solutions by eliminating a need for multiple trials for searching optimal training parameters.


\medskip
\small

\bibliography{bibtex}

\clearpage
\appendixpage

\begin{appendices}
\section{Dropout}
\label{appendix:dropout}
\numberwithin{equation}{section}
\setcounter{equation}{0}
 
To model the degradation of the information in features due to dropout we consider the information content of a sign of a feature with one-bit representation of the feature that corresponds to its sign only with values $X=0,1$. We consider the dropout layer representation that is completely equivalent to a standard definition.
In this model the dropout layer is a gate unit that transfers the feature. The gate is controlled by a random bit - the dropout noise. When the feature is positive the feature input to the dropout layer is one $X=1$, otherwise the input bit is zero $X=0$. When the random dropout bit is one the gate transfers the feature as is and when the random dropout bit is zero the feature is transferred as zero. 

For input probability distribution $P(X)$ the output distribution of the dropout layer $P(Y|X)$ will be

\begin{equation}\label{a:dropout:distr}
P(Y|X=0) = \delta_{Y,0},\; P(Y|X=1) = r\delta_{Y,0}+(1-r)\delta_{Y,1}. 
\end{equation}

Here r is the dropout rate from the interval $0\leqslant r <1$.

Due to dropout the information content in the features is reduced. The information measure of the output of the dropout layer is given by the mutual information $I(X;Y)$

\begin{equation}
I(X;Y) = \sum_{X,Y}{p(X)p(Y|X)\log_2{\left(\frac{p(Y|X)}{p(Y)}\right)}}.
\end{equation}

Using the distributions in the eq.\eqref{a:dropout:distr} we can find

\begin{equation}\label{a:dropout:mutual}
\begin{split}
I(X;Y|r) = & -\left\{P(X=0)+rP(X=1)\right\}\log_2\left\{P(X=0)+rP(X=1)\right\}\\
& - P(X=1)\left\{r\log_2{\frac{1}{r}}+(1-r)\log_2P(X=1)\right\}.
\end{split}
\end{equation}

When $r=1$ the information is completely lost $I=0$. When $r=0$ the mutual information in the eq.\eqref{a:dropout:mutual} is equal to the information in $X$

\begin{equation}
I(X;Y|r=0) = -P(X=0)\log_2{P(x=0)}-P(X=1)\log_2{P(X=1)}.
\end{equation}

The ratio $R=I(X;Y|r)/I(X;Y|r=0)$ gives an effective reduction factor for the amount of information in a feature due to dropout noise. The reduction factor $R=1$ when $r=0$ and $R=0$ when $r=1$. Essentially this factor is reducing the effective dimensionality of the features due to dropout. Then using the dropout with different rates we can continuously control the effective number of feature dimensions.

\section{The integral factorization error}
\label{appendix:integral}

The factorization of the integral in the eq.\eqref{int:factor} results in the error that is defined by the right side of the Cauchy-Schwarz inequality in the eq.\eqref{cauchy-schwarz}. 
To compute the error we have to expand the $\mathbf{A}$ and $\mathbf{B}$ to the power series by $\epsilon$ up to the second order $O(\epsilon^2)$ and compute corresponding expansion of the product of variances $\mathbf{Var(A)}, \mathbf{Var(B)}$ up to the second-order of $\epsilon$. Remember that $\mathbf{\langle A \rangle_{H_1}} = \int{\mathrm{dw}P(w|\mathbf{H_1}) A(w)}$. So the mean of $\mathbf{A}$ by $P(w|\mathbf{H_1})$ is 

\begin{equation}
\mathbf{Mean(A)_{H_1}} = \mathbf{Mean}\left(\frac{P(w|\mathbf{H_0})}
{P(w|\mathbf{H_1})}e^{-\epsilon L}\right)_{\mathbf{H_1}} = \mathbf{Mean}\left( e^{-\epsilon L} \right)_{\mathbf{H_0}}.
\end{equation} 

Then the variance is

\begin{equation}
\begin{split}
\mathbf{Var(A)} = & \langle\left(\frac{P(w|\mathbf{H_0})}
{P(w|\mathbf{H_1})}e^{-\epsilon L}\right)^2\rangle_{\mathbf{H_1}} - 
\left(\mathbf{Mean(A)_{H_1}}\right)^2 = \\
& \langle\left(\frac{P(w|\mathbf{H_0})}
{P(w|\mathbf{H_1})}e^{-2\epsilon L}\right)\rangle_{\mathbf{H_0}} -
\langle e^{-\epsilon L}\rangle_{\mathbf{H_0}}^2.
\end{split}
\end{equation}

The new hyper-parameter point $\mathbf{H_1=H_0 + \Delta H}$, where $\Delta H$ has an order of $O(\epsilon)$. The expansion of the variance of $\mathbf{A}$ by $\epsilon$ up to the second order gives

\begin{equation}
\mathbf{Var(A)_{H_1}} = \epsilon^2 \mathbf{Var}(L)_{\mathbf{H_0}} + 
2\epsilon \langle L \mathbf{\Delta H}\frac{\partial\log{P(w|\mathbf{H_0})}}{\partial\mathbf{H}}\rangle_{\mathbf{H_0}} + 
\langle \left( \mathbf{\Delta H}\frac{\partial\log{P(w|\mathbf{H_0})}}{\partial\mathbf{H}}\right)^2\rangle_{\mathbf{H_0}}.
\end{equation}

The variance of $\mathbf{B}$ at $\mathbf{H_1}$ is

\begin{equation}
\mathbf{Var(B)_{H_1}} = 
\langle e^{-2(1-\epsilon)L}\rangle_{\mathbf{H_1}} - 
\langle e^{-(1-\epsilon)L}\rangle_{\mathbf{H_1}}^2.
\end{equation}

The leading term in the variance of $\mathbf{B}$ is $O(1)$ by $\epsilon$ and does not depend on $\mathbf{\Delta H}$. Then we can take into account that $e^{-L}$ is a bounded random variable and so its variance is bounded by a constant $O(1)$, see details in  \citep{popoviciu1935equations}, \citep{doi:10.1080/00029890.2000.12005203}.

Now we can reduce the integral factorization error by minimizing $\mathbf{Var(A)_{H_1}}$ by $\mathbf{\Delta H}$, which immediately gives the update rule in the eq.\eqref{int:update}. Using the $\mathbf{\Delta H}$ from the eq.\eqref{int:update} we get the minimal value of the $\mathbf{Var(A)_{H_1}}$

\begin{equation}\label{a:int:var:A}
\min_{\mathbf{H_1}}\{\mathbf{Var(A)_{H_1}}\} = \epsilon^2 \mathbf{Var}(L)_{\mathbf{H_0}} - \epsilon^2 
\frac{\partial \langle L  \rangle_{\mathbf{H_0}}}{\partial\mathbf{H}}
\left\langle \left(\frac{\partial\log{P(w|\mathbf{H_0})}}{\partial\mathbf{H}} \right)^2\right\rangle_{\mathbf{H_0}}^{-1}
\frac{\partial \langle L  \rangle_{\mathbf{H_0}}}{\partial\mathbf{H}}.
\end{equation}

Because total loss $L$ is proportional to a number of samples $N$ the second term in the eq.\eqref{a:int:var:A} seems to be quadratic by $N$. However, this is not the case we would like to consider. We would like to select hyper-parameters of the prior to make the prior $P(w|\mathbf{H})$ having a sharp peak by $w$ to be important in the integral \eqref{int:int} relative to losses and that requires the $\log{P(w|\mathbf{H})}$ to be at least of order $O(N)$ near maximum. Then the second term in the eq.\eqref{a:int:var:A} is $O(1)$ by $N$. The first term is the variance of a large sum of $N$ individual sample losses. Considering that the samples are independent and random the variance of $\mathbf{A}$ is proportional to $N$.

Finally, the factorization error is a square root of the product of variances of $\mathbf{A}$ and $\mathbf{B}$ which gives the integral  factorization error estimate for one $\epsilon$-step to be $O(\epsilon \sqrt{N})$. After $T$ steps the error accumulates to a factor $\exp(O(\sqrt{N}))$ because $T\epsilon=1$.

\section{BSGD: Approximating second gradient of loss}
\label{appendix:bsgd}

While computing of the first gradient of a sample loss w.r.t. weights is a standard operation in typical neural network frameworks which is   implemented via backpropagation algorithm the computing of the second gradient of a sample loss function usually is not available.

However, availability of the second gradient is not a main problem.
Because the loss of a sample is defined as a negative log of a model probability $l(x,w)=-\log{P(x|w)}$ the second gradient of a loss can be represented as a sum of two parts 

\begin{equation}\label{a:bsgd:second_grad}
\frac{\partial^2 l}{\partial w^2} = \left(\frac{\partial l}{\partial w}\right)^2 - \frac{1}{P(x|w)}\frac{\partial^2 P(x|w)}{\partial w^2}.
\end{equation}

The second term on the right side of the equation could have large negative values from sharp minima of the model probability that may destabilize a convergence of BSGD in the eq.\eqref{bsgd:final}. To stabilize the iterations of the variances we need to modify that second term. After $t$ iterations the variance will be

\begin{equation}
\frac{1}{\sigma_{i,t}^2} = \frac{1}{\sigma_{i,0}^2} + \sum_t{\frac{1}{T}\frac{\partial^2{l_t(w)}}{\partial{w_{i,t}^2}}}
\end{equation}

We can notice that the total contribution from the term with the second gradient of the probability in eq.\eqref{a:bsgd:second_grad} for all updates of the variance in one epoch in the eq.\eqref{bsgd:final} accumulates to the  following value

\begin{equation}
\int{\mathrm{dw}P(w|\mathbf{H})\;\frac{1}{N}\sum_{n=1}^{N}{\frac{1}{P(x_n|w)} \frac{\partial^2 P(x_n|w)}{\partial w^2}}}.
\end{equation}

The sum over samples is an approximation of the average with an unknown true distribution $P(x)$

\begin{equation}\label{a:bsgd:int}
\frac{1}{N}\sum_{n=1}^{N}{\frac{1}{P(x_n|w)} \frac{\partial^2 P(x_n|w)}{\partial w^2}} \approx \int{\mathrm{dx}\frac{P(x)}{P(x|w)} \frac{\partial^2 P(x|w)}{\partial w^2}}.
\end{equation}

The sampling error of replacing the sum with the integral estimate in the eq.\eqref{a:bsgd:int} is $O(1/\sqrt{N})\ll 1$. For typically used large $N$ the sampling error is small.
 
Now we can see that if the model distribution $P(x|w)$ is close or rather proportional up to constant in the area of known samples to the true distribution $P(x)$, which we could expect if the model converges, then the integral over $x$ is diminishing as well as an overall contribution of the term with second gradient of the model probability to the cumulative sum.

The solution to avoid the destabilizing spikes from the second term on the right side in the eq.\eqref{a:bsgd:second_grad} could be a smoothing of the term compatible with its overall disappearance from the complete epoch sum, however just omitting the second term is found to be a good solution, so in the definition of the BSGD we will approximate the second gradient of a loss with a square of the first gradient of the loss.

\section{Network definition}
\label{appendix:net}

The neural network used for experiments is a convolutional network with a feed-forward architecture.

\begin{figure}[h]
\begin{center}
\includegraphics[scale=0.3]{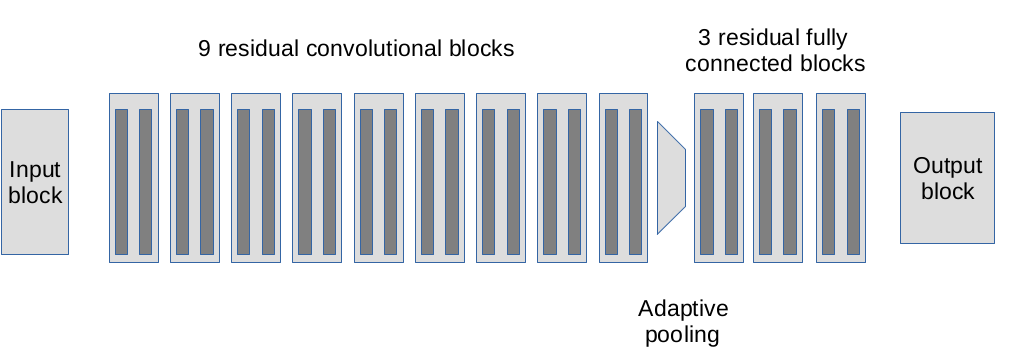}
\end{center}
\caption{Neural network diagram.}
\label{a:fig:net}
\end{figure}

The network diagram is in the Figure \ref{a:fig:net}.

The network consists of four large blocks: input block, feature extraction block, prediction block and output block. 

The input block consists of a single convolutional layer with a kernel size 5x5 that converts one channel input to 100 channels followed by ReLU activations. 

Feature extraction block has 9 residual blocks with 2 convolutional layers per each block with a kernel size 3x3 and ReLU activations.

Feature extraction block is followed by an adaptive pooling layer.

The following prediction block has 3 residual blocks each with 2 fully connected linear layers and ReLU activations.

The output block is a linear layer that converts 100 features to 10.

The cross-entropy loss is implemented via log-softmax function.

\end{appendices}

\end{document}